\documentclass[10pt,twocolumn,letterpaper]{article}

\usepackage[pagenumbers]{cvpr} 

\usepackage{graphicx}
\usepackage{amsmath}
\usepackage{amssymb}
\usepackage{booktabs}
\usepackage{multirow}
\usepackage{enumitem}
\usepackage[table,xcdraw]{xcolor}
\usepackage{comment}
\usepackage{arydshln}
\usepackage{stfloats}
\usepackage[pagebackref,breaklinks,colorlinks]{hyperref}

\usepackage[capitalize]{cleveref}
\crefname{section}{Sec.}{Secs.}
\Crefname{section}{Section}{Sections}
\Crefname{table}{Table}{Tables}
\crefname{table}{Tab.}{Tabs.}

\def\cvprPaperID{6160} 
\def\confName{CVPR}
\def\confYear{2023}

\begin{document}

\title{PillarNeXt: Rethinking Network Designs for 3D Object Detection\\ in LiDAR Point Clouds}

\author{Jinyu Li \qquad Chenxu Luo \qquad Xiaodong Yang\thanks{Corresponding author \texttt{xiaodong@qcraft.ai}}\\
QCraft\\
}
\maketitle

\begin{abstract}
In order to deal with the sparse and unstructured raw point clouds, LiDAR based 3D object detection research mostly focuses on designing dedicated local point aggregators for fine-grained geometrical modeling. In this paper, we revisit the local point aggregators from the perspective of allocating computational resources. We find that the simplest pillar based models perform surprisingly well considering both accuracy and latency. Additionally, we show that minimal adaptions from the success of 2D object detection, such as enlarging receptive field, significantly boost the performance. Extensive experiments reveal that our pillar based networks with modernized designs in terms of architecture and training render the state-of-the-art performance on the two popular benchmarks: Waymo Open Dataset and nuScenes. Our results challenge the common intuition that the detailed geometry modeling is essential to achieve high performance for 3D object detection.
\end{abstract}

\section{Introduction}
\label{sec:intro}
3D object detection in LiDAR point clouds is an essential task in an autonomous driving system, as it provides crucial information for subsequent onboard modules, ranging from perception~\cite{simtrack, pillar-motion}, prediction~\cite{prophnet, wayformer} to planning~\cite{plan1, plan2}. There have been extensive research efforts on developing sophisticated networks that are specifically designed to cope with point clouds in this field~\cite{lang2019pointpillars, shi2022pillarnet, shi2021pv, zhou2018voxelnet, wang2020pillar}. 

\begin{figure}[t]
  \centering
   \includegraphics[width=\linewidth]{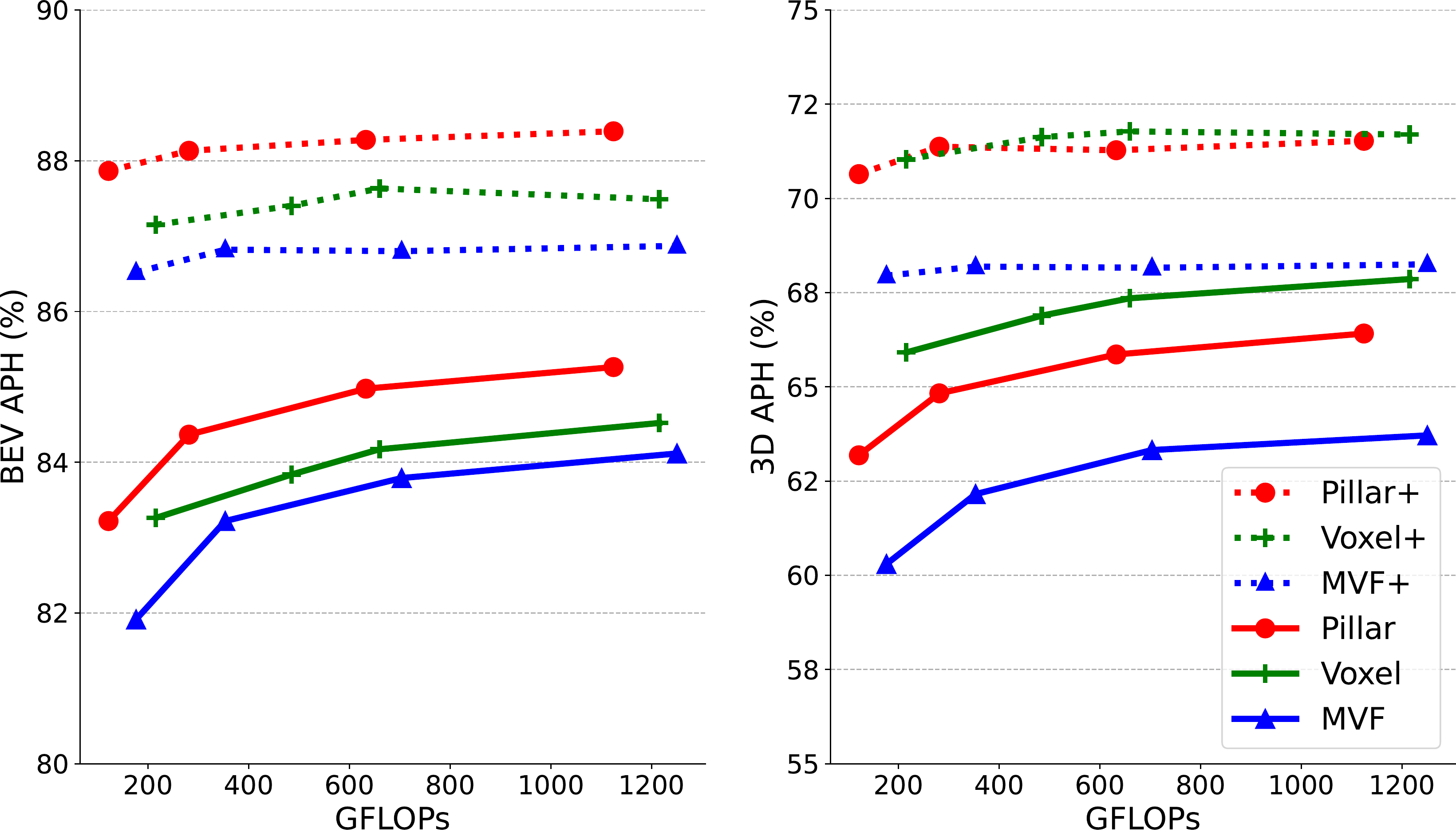}
   \vspace{-7mm}
   \caption{Overview of pillar, voxel and multi-view fusion (MVF) based 3D object detection networks under different GFLOPs. The dash lines denote the enhanced versions of corresponding models (+). We report the L2 BEV and 3D APH of vehicle on the validation set of Waymo Open Dataset.}  
   \vspace{-2mm}
   \label{fig:teaser}
\end{figure}

Due to the sparse and irregular nature of point clouds, most existing works adopt the grid based methods, which convert point clouds into regular grids, such as pillar~\cite{lang2019pointpillars}, voxel~\cite{zhou2018voxelnet} and range view~\cite{meyer2019lasernet}, such that regular operators can be applied. 
However, it is a common belief that the grid based methods (especially for pillar) inevitably induce information loss, leading to inferior results, in particular for small objects (e.g., pedestrians). Recent research~\cite{shi2021pv} proposes the hybrid design for fine-grained geometrical modeling to combine the point and gird based representations. We name all the above operators as local point aggregators since they aim to aggregate point features in a certain neighborhood. We observe that the current mainstream of 3D object detection is to develop more specialized operators for point clouds, while leaving the network architecture almost unexplored. Most existing works are still built upon the very original architectures of SECOND~\cite{yan2018second} or PointPillars~\cite{lang2019pointpillars}, which lacks of modernized designs.

Meanwhile, in a closely related area, 2D object detection in images has achieved remarkable progress, which can be largely attributed to the advances in architecture and training. Among them, the powerful backbones (e.g., ResNet~\cite{he2016deep} and Swin Transformers~\cite{liu2021swin}), the effective necks (e.g., BiFPN~\cite{tan2020efficientdet} and YOLOF~\cite{yolof}), and the improved training (e.g., bag of freebies~\cite{freebies, yolov4}) are in particular notable. Therefore, the research focus of 2D object detection is largely different from that of 3D object detection.

In light of the aforementioned observations, we rethink what should be the focus for 3D object detection in LiDAR point clouds. Specifically, we revisit the two fundamental issues in designing a 3D object detection model: the local point aggregator and the network architecture. 

First, we compare local point aggregators from a new perspective, i.e., the computational budget. A fine-grained aggregator usually demands more extensive computing resources than the coarse one. For instance, the voxel based aggregator employs 3D convolutions, which require more network parameters and run much slower than 2D convolutions. This raises a question of how to effectively allocate the computational budget or network capacity. Should we spend the resources on fine-grained structures or assign them to coarse grids? Surprisingly, as shown in Figure~\ref{fig:teaser}, when training under a comparable budget and with an enhanced strategy, 
a simpler pillar based model can achieve superior or on par performance with the voxel based model, even for small objects such as pedestrians, while significantly outperform the multi-view fusion based model. 
This challenges the actual performance gain and the necessity of fine-grained local 3D structures. Our finding is also consistent with the recent works~\cite{liu2020closer, qian2022pointnext} in general point cloud analysis, demonstrating that different local point aggregators perform similarly under strong networks. 

Second, for the network architecture, we do not aim to propose any domain specific designs for point clouds, instead, we make minimal adaptations from the success of 2D object detection and show that they outperform most of existing networks with specialized architectures for point clouds. For instance, one key finding is that enlarging receptive field properly brings significant improvement. Unlike previous works~\cite{zhou2018voxelnet, shi2022pillarnet} relying on multi-scale feature fusion, we show that a single scale at final stage with sufficient receptive field obtains better performance. Such promising results suggest that 3D object detection can inherit the successful practices well developed in 2D domain.

Levaraging on the findings above, we propose a pillar based network, dubbed as \textbf{PillarNeXt}, which leads to the state-of-the-art results on two popular benchmarks~\cite{Sun_2020_CVPR, caesar2020nuscenes}. Our approach is simple yet effective, and enjoys strong scalability and generalizability. We develop a series of networks with different trade-offs between detection accuracy and inference latency through tuning the number of network parameters, 
which can be used for both on-board~\cite{lang2019pointpillars} and off-board~\cite{qi2021offboard} applications in autonomous driving.

Our main contributions can be summarized as follows. (1) To our knowledge, this is the first work that compares different local point aggregators (pillar, voxel and multi-view fusion) from the perspective of computational budget allocation. Our findings challenge the common belief by showing that pillar can achieve comparable 3D mAP and better bird's eye view (BEV) mAP compared to voxel, and substantially outperform multi-view fusion in both 3D and BEV mAP. (2) Inspired by the success of 2D object detection, we find that enlarging receptive field is crucial for 3D object detection. With minimal adaptions, our detectors outperform existing methods with sophisticated designs for point clouds. (3) Our networks with appropriate training achieve superior results on two large-scale benchmarks. We hope our models and related findings can serve as a strong and scalable baseline for future research in this community. Our code and model will be made available at \url{https://github.com/qcraftai/pillarnext}.

\section{Related Work}
\noindent\textbf{LiDAR based 3D Object Detection.} 
Existing methods can be roughly categorized into point, grid and hybrid based representations, according to the local point aggregators. As a point based method, PointRCNN~\cite{shi2019pointrcnn} generates proposals using~\cite{qi2017pointnet++} and then refines each proposal by RoI pooling. However, conducting neighboring point aggregation in such methods is extremely expensive, so they are not feasible to handle large-scale point clouds in autonomous driving. On the other hand, the grid based methods discretize point clouds into structured grids, where 2D or 3D convolutions can be applied. In~\cite{zhou2018voxelnet} VoxelNet partitions a 3D space into voxels and aggregates point features inside each voxel, then dense 3D convolutions are used for context modeling. SECOND~\cite{yan2018second} improves the efficiency by introducing sparse 3D convolutions. PointPillars~\cite{lang2019pointpillars} organizes point clouds as vertical columns and adopts 2D convolutions. Another grid based representation is the range view~\cite{meyer2019lasernet} that can be also efficiently processed by 2D convolutions. The multi-view fusion methods~\cite{zhou2020end,wang2020pillar} take advantage of both pillar/voxel and range view based representations. 

In spite of efficiency, it is commonly and intuitively believed that the grid based methods induce fine-grained information loss. Therefore, the hybrid methods are proposed to incorporate point features into grid representations~\cite{liu2021pvnas,shi2021pv}. In this work, we instead focus on the basic network architecture and associated training, and show that the fine-grained local geometrical modeling is overestimated.

\noindent\textbf{Feature Fusion and Receptive Field.} 
The multi-scale feature fusion starts from feature pyramid network (FPN)~\cite{lin2017feature} that aggregates hierarchical features in a top-down manner. It is widely used in 2D object detection to combine high-level semantics with low-level spatial cues. In~\cite{liu2018path}, PANet further points out the bottom-up fusion is also important. Both of them perform fusion by adding up feature maps directly. BiFPN~\cite{tan2020efficientdet} shows that features from different scales contribute unequally, and adopts learnable weights to adjust the importance. As an interrelated factor of feature fusion, receptive field is also broadly studied and verified in 2D detection and segmentation. In~\cite{Chen_2018_PAMI}, the atrous spatial pyramid pooling (ASPP) is proposed to sample features with multiple effective receptive fields. TridentNet~\cite{Li_2019_ICCV} applies three convolutions with different dilation factors to make receptive field range to match with object scale range. A similar strategy is also introduced in YOLOF~\cite{yolof} that employs a dilated residual block for enlarging receptive field and meanwhile keeping original receptive field. 

Although these techniques regarding feature fusion and receptive field have been extensively adopted in 2D domain, they are hardly discussed in 3D domain. Most existing networks in this field still follow or modify based on the architecture of VoxelNet~\cite{zhou2018voxelnet}. In this work, we aim to integrate the up-to-date designs into 3D object detection networks.

\begin{figure*}[t]
  \centering
   \includegraphics[width=0.86\linewidth]{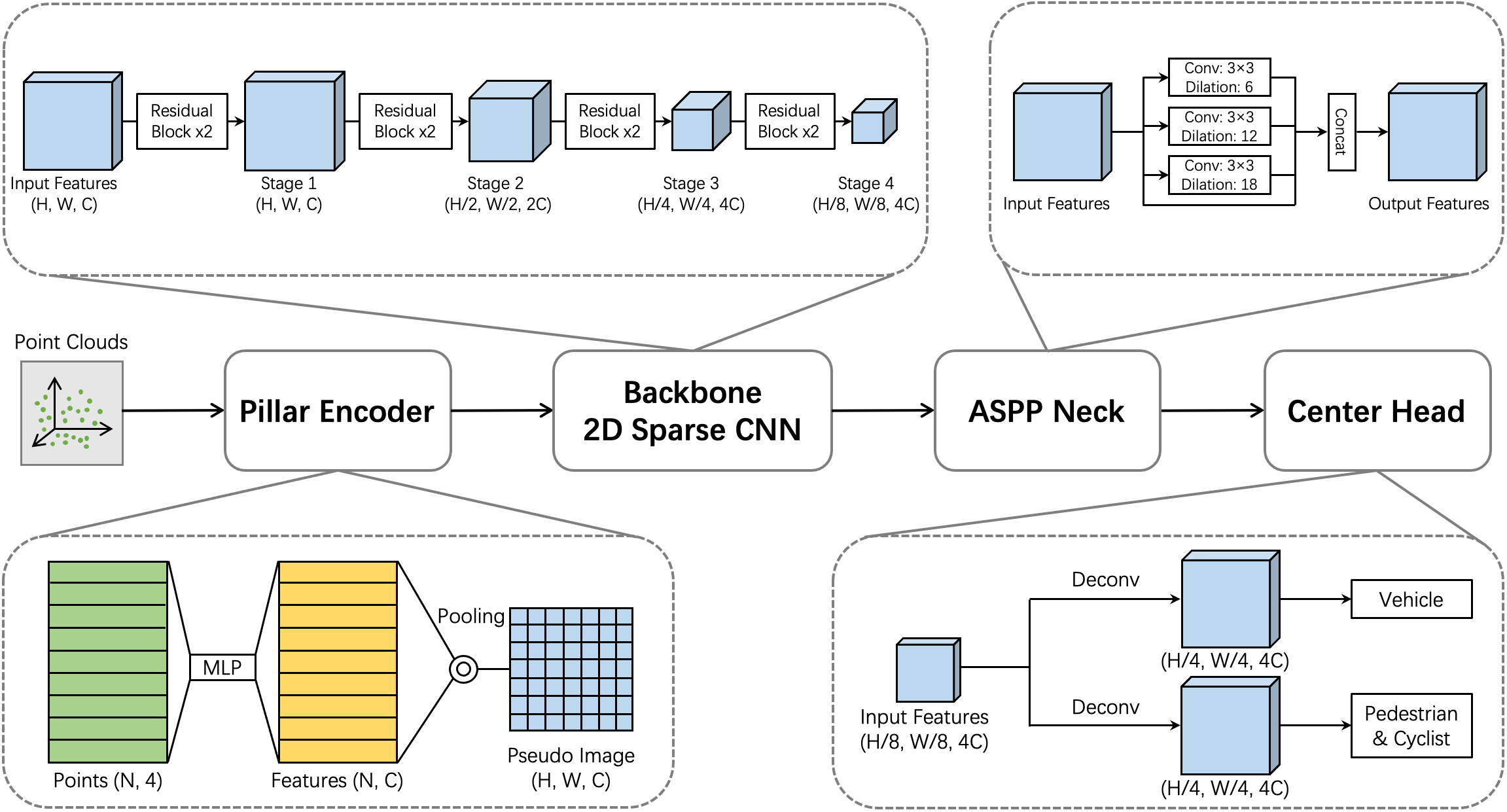}
   \caption{A schematic overview of the network architecture of the proposed PillarNeXt. Our model takes raw point clouds as input and relies on a simple pillar encoder, which consists of MLPs and max pooling to convert point clouds into a pseudo image. We then apply ResNet-18 with sparse convolutions as the backbone, and adopt ASPP based neck to enlarge receptive field. After that, we upsample the feature maps to yield more detailed representations, and use the center based multi-group head to produce the detection output.} 
   \vspace{-1mm}
   \label{fig:sup-arch}
\end{figure*}

\noindent\textbf{Model Scaling.} 
This aspect has been well studied in image domain, including classification, detection and segmentation tasks. It is observed that jointly increasing the depth, width and resolution improves the accuracy. EfficientNet in~\cite{tan2019efficientnet} proposes a compound scaling rule for classification, and this rule is later on extended to object detection~\cite{tan2020efficientdet, dollar2021fast}. It is a general consensus that the model capacity affects the model performance. Therefore, the comparison of different methods should be under the consideration of model capacity in order to receive sound conclusions.

There is only one work~\cite{wang2022cost} that studies model scaling in 3D object detection, to the best of our knowledge. It scales the depth, width and resolution of SECOND~\cite{yan2018second} to find the importance of each component. It however only focuses on a single type of model, while we systematically compare different local point aggregators under similar computational budgets across a wide range of model scales.

\section{Network Architecture Overview}
We focus on the grid based models due to the runtime efficiency and proximity to 2D object detection. 
Typically, a grid based network is composed of (i) a grid encoder to convert raw point clouds into structured feature maps, (ii) a backbone for general feature extraction, (iii) a neck for multi-scale feature fusion, and (iv) a detection head for the task-specific output. Existing networks often couple all these components together. In this section, we have them decoupled and review each part briefly. 
\subsection{Grid Encoder} 
\label{grid-encoder}
A grid encoder is used to discretize point clouds into structured grids, and then convert and aggregate the points within each grid into the preliminary feature representation. In this work, we target at the following three grid encoders. 

\begin{itemize}[noitemsep,topsep=0pt]
\item \textbf{Pillar:} A pillar based grid encoder arranges points in vertical columns, and applies multilayer perceptrons (MLPs) followed by max pooling to extract pillar features, which are represented as a pseudo image~\cite{lang2019pointpillars}. 
\item \textbf{Voxel:} Similar to pillar, the voxel based grid encoder organizes points in voxels and obtains corresponding features~\cite{zhou2018voxelnet}. Compared with pillar, the voxel encoder preserves details along the height dimension.  
\item \textbf{Multi-View Fusion (MVF):} A MVF based grid encoder combines the pillar/voxel and range view based representations. Here we follow~\cite{wang2020pillar} to incorporate a pillar encoder with a cylindrical view based encoder that groups points in the cylindrical coordinates.     
\end{itemize}

\begin{table*}[t]
\centering
\begin{tabular}{c|lccc|cc|cc}
\hline
\multirow{2}{*}{Model} & \multicolumn{1}{c}{\multirow{2}{*}{Channels}} & \multirow{2}{*}{\#Params (M)} & \multirow{2}{*}{FLOPs (G)} & \multirow{2}{*}{Latency (ms)}& \multicolumn{2}{c|}{Vehicle} & \multicolumn{2}{c}{Pedestrian} \\ \cline{6-9} 
                        &                          &        &      &                & 3D           & BEV           & 3D             & BEV           \\ \hline
  Pillar-T       &   [32, 64, 128, 128]                        &       1.65    & 70                   &          52    &  \textbf{62.03}        &     \textbf{82.26}   &     \textbf{67.63}      &     \textbf{75.76}       \\  
  MVF-T & [32, 64, 128, 128] & 3.44 & 78 & 137 & 59.16 & 81.33 & 64.10 & 73.42 \\
   \hline 
  Pillar-S & [42, 84, 168, 168]  & 2.83  & 121 & 79 & 63.18 & \textbf{83.22} & 68.12 & \textbf{76.37} \\
 Voxel-S      &     [12, 24, 48, 96]     &       1.53          & 121             &      169        &     \textbf{64.67}      &    82.45      &     \textbf{69.10}      &     76.29     \\ 
 MVF-S & [44, 88, 176, 176] & 6.38 & 148 & 186 & 61.06 & 82.51 & 65.15 &74.24  \\
  \hline 
  Pillar-B & [64, 128, 256, 256]  & 6.53 & 281 &  103 & 64.83 & \textbf{84.37 }&  69.04 & \textbf{76.96}\\ 
  Voxel-B & [18, 36, 72, 144]  & 3.42 & 272 & 226 &  \textbf{66.00} & 83.32 &  \textbf{69.45} & 76.38\\
  MVF-B & [68, 136, 272, 272] & 15.02 & 353 & 291 & 62.15 & 83.22 & 66.12 & 75.00  \\ \hline
  Pillar-L & [96, 192, 384, 384] & 14.63 & 632 & 194 & 65.86 & \textbf{84.98} & 68.42 & 76.67 \\ 
  Voxel-L & [28, 56, 112, 224]  &8.27 &  660  & 299 &  \textbf{67.35} & 84.17 & \textbf{70.47} & \textbf{77.44} \\ 
 MVF-L & [96, 192, 384, 384] & 29.67 & 704 & 390 & 63.32 & 83.79 & 66.87 &  75.34   \\ \hline
\rowcolor[HTML]{EFEFEF} Pillar-S+  & [42, 84, 168, 168]  & 2.83  & 121 & 79 &  \textbf{70.64} & \textbf{87.86} & 73.48 & 79.95 \\  
\rowcolor[HTML]{EFEFEF} Voxel-S+  & [12, 24, 48, 96]  & 1.53  & 121 & 169 &  70.61 & 86.83 & \textbf{74.26} & \textbf{80.37} \\ 
\rowcolor[HTML]{EFEFEF} MVF-S+ & [44, 88, 176, 176] & 6.38 & 148 & 186 & 67.46 & 86.52 & 70.87 & 78.18  \\ \hline
\rowcolor[HTML]{EFEFEF} Pillar-B+ & [64, 128, 256, 256] & 6.53 & 281 & 103 &   \textbf{71.37} & \textbf{88.13} & 73.93 & 80.28 \\
\rowcolor[HTML]{EFEFEF} Voxel-B+ & [18, 36, 72, 144]  & 3.42 & 272 & 226  &  71.36 & 87.33 & \textbf{74.76} & \textbf{80.74} \\
\rowcolor[HTML]{EFEFEF} MVF-B+ & [68, 136, 272, 272] & 15.02 & 353 & 291  &  68.19 & 86.82 & 71.42 & 78.51 \\ \hline 
\end{tabular}
\caption{Comparison of the pillar, voxel and MVF based networks with model scales from tiny, small, base to large. We report the L2 3D and BEV APH on vehicle and pedestrian on the validation set of WOD. Groups 1 and 2 correspond to the regular and enhanced versions.}  
\label{tab:capacity}
\end{table*}

\subsection{Backbone and Neck}
A backbone performs further feature abstraction based on the preliminary features extracted by the grid encoder. For fair comparisons, we utilize ResNet-18 as the backbone, since it is commonly used in the previous works~\cite{zhu2019class,yin2021center,shi2022pillarnet}. Specifically, we make use of sparse 2D convolutions in the backbone with the pillar or MVF based encoder, and sparse 3D convolutions in the backbone with the voxel based encoder. A neck can be then utilized to aggregate features from the backbone for enlarging receptive field and fusing multi-scale context. However, how to design an effective neck has not been well explored for object detection in point clouds compared to in images. We aim to close this gap by integrating the advanced neck designs from 2D object detection, such as BiFPN~\cite{tan2020efficientdet} using improved multi-level feature fusion or ASPP~\cite{Chen_2018_PAMI} using convolutions with multiple dilated rates on a single feature level, into the model architectures of 3D object detection.

\subsection{Detection Head}
In the pioneering works of SECOND~\cite{yan2018second} and PointPillars~\cite{lang2019pointpillars}, the anchor based detection head is employed to pre-define the axis-aligned anchors at each location on the input feature maps to head. CenterPoint~\cite{yin2021center} instead represents each object by its center point, and predicts a centerness heatmap where the regression of bounding box is realized at each center location. 
Due to its simplicity and superior performance, we adopt the center based detection head in all our networks. We show a set of simple modifications in head, such as feature upsampling, multi-grouping and IoU branch, improve the performance notably.  
\section{Experiments}

In this section, we start from introducing the experimental setup including datasets and implementation details. We then perform comprehensive network design studies on each component in a 3D object detection model. In the end, we present extensive comparisons with the state-of-the-art methods on the two popular benchmarks. 

\subsection{Experimental Setup}
We conduct experiments on two large-scale autonomous driving benchmarks: Waymo Open Dataset (WOD)~\cite{Sun_2020_CVPR} and nuScenes~\cite{caesar2020nuscenes}. \textbf{WOD} consists of 798 sequences (160K frames) for training and 202 sequences (40K frames) for validation, which are captured with 5 LiDARs at 10Hz. Following the official protocol, we use the average precision (AP) and average precision weighted by heading (APH) as the evaluation metrics. We break down the performance into two difficulty levels, L1 and L2, where the former evaluates objects with at least 5 points and the latter covers all objects with at least one point. We set the IoU thresholds for vehicle, pedestrian and cyclist to 0.7, 0.5 and 0.5. In addition to 3D AP/APH, we also report the results under BEV. \textbf{nuScenes} contains 1,000 scenes of roughly 20 seconds each, captured by a 32-beam LiDAR at 20Hz. Annotations are available on keyframes at 2Hz. We follow the official evaluation metrics by averaging over 10 classes under mean average precision (mAP) and nuScenes detection score (NDS), and the latter is a weighted average of mAP as well as ATE, ASE, AOE, AVE and AAE that are used to respectively measure the translation, scale, orientation, velocity and attribute related errors.

\begin{table*}[t]
\centering
\begin{tabular}{c|cc|cc|cc|cc}
\hline
\multirow{2}{*}{Method}  & \multicolumn{2}{c|}{Vehicle L1} & \multicolumn{2}{c|}{Vehicle L2} & \multicolumn{2}{c|}{Pedestrian L1} & \multicolumn{2}{c}{Pedestrian L2} \\ \cline{2-9} 
         &  AP & APH & AP & APH & AP & APH & AP & APH  \\ \hline
Neck of PillarNet~\cite{shi2022pillarnet}  & 91.39 & 90.58  &  84.54 & 83.72  & \textbf{87.90}  &  \textbf{83.02} & 81.93  & 77.20    \\
FPN~\cite{lin2017feature}  &92.17 & 91.35 & 85.96 & 85.13 & 87.88 & 82.91 & 82.05 & 77.23\\
BiFPN~\cite{tan2020efficientdet} & 92.71 & 91.90 & 86.92 & 86.09 & 87.86 & 82.88 & 82.05 & 77.23 \\ 
\hline
Plain & 91.01 & 90.19 & 83.86 & 83.04 & 87.59 & 82.61 & 81.52 & 76.71 \\
Dilated Block~\cite{yolof}  & 92.70 & 91.90  &  86.61 & 85.79  & 87.84  & 82.91  &  \textbf{82.09} & \textbf{77.29}    \\
ASPP~\cite{Chen_2018_PAMI} & \textbf{92.77} & \textbf{91.94} & \textbf{86.99} & \textbf{86.14} & 87.74& 82.85 & 82.00 & 77.26 \\  \hline
\end{tabular}
\vspace{-2mm}
\caption{Comparison of different neck modules integrated in our networks. Groups 1 and 2 correspond to the multi-scale and sing-scale necks, respectively. We report the L1 and L2 BEV AP and APH for vehicle and pedestrian on the validation set of WOD.} 
\vspace{-1mm}
\label{tab:neck}
\end{table*}

We implement our networks in PyTorch. All models are trained by using AdamW~\cite{adamW} as the optimizer and under the one-cycle~\cite{onecycle} learning rate schedule. 
For \textbf{WOD}, the detection range is set to [-76.8m, 76.8m] horizontally and [-2m, 4m] vertically. Our pillar size is 0.075m in x/y-axis (and 0.15m in z-axis for voxel based). 
For MVF based models, we keep the same pillar size and use [1.8$^{\circ}$, 0.2m] for yaw and z-axis in the cylindrical view. We train each model for 12 epochs and take 3 frames as input unless otherwise specified. For inference, we use the non-maximum suppression (NMS) thresholds of 0.7, 0.2 and 0.25 for vehicle, pedestrian and cyclist. As for \textbf{nuScenes}, we take 10 frames as input, and set detection range to [-50.4m, 50.4m] horizontally and [-5m, 3m] vertically (0.2m in z-axis for voxel based). For inference, we use NMS threshold of 0.2 for all classes. Other settings are the same as WOD. We report the inference latency on a single NVIDIA TITAN RTX GPU. 
More implementation details can be found in Appendix. 

\begin{figure}[t]
  \centering
   \includegraphics[width=0.8\linewidth]{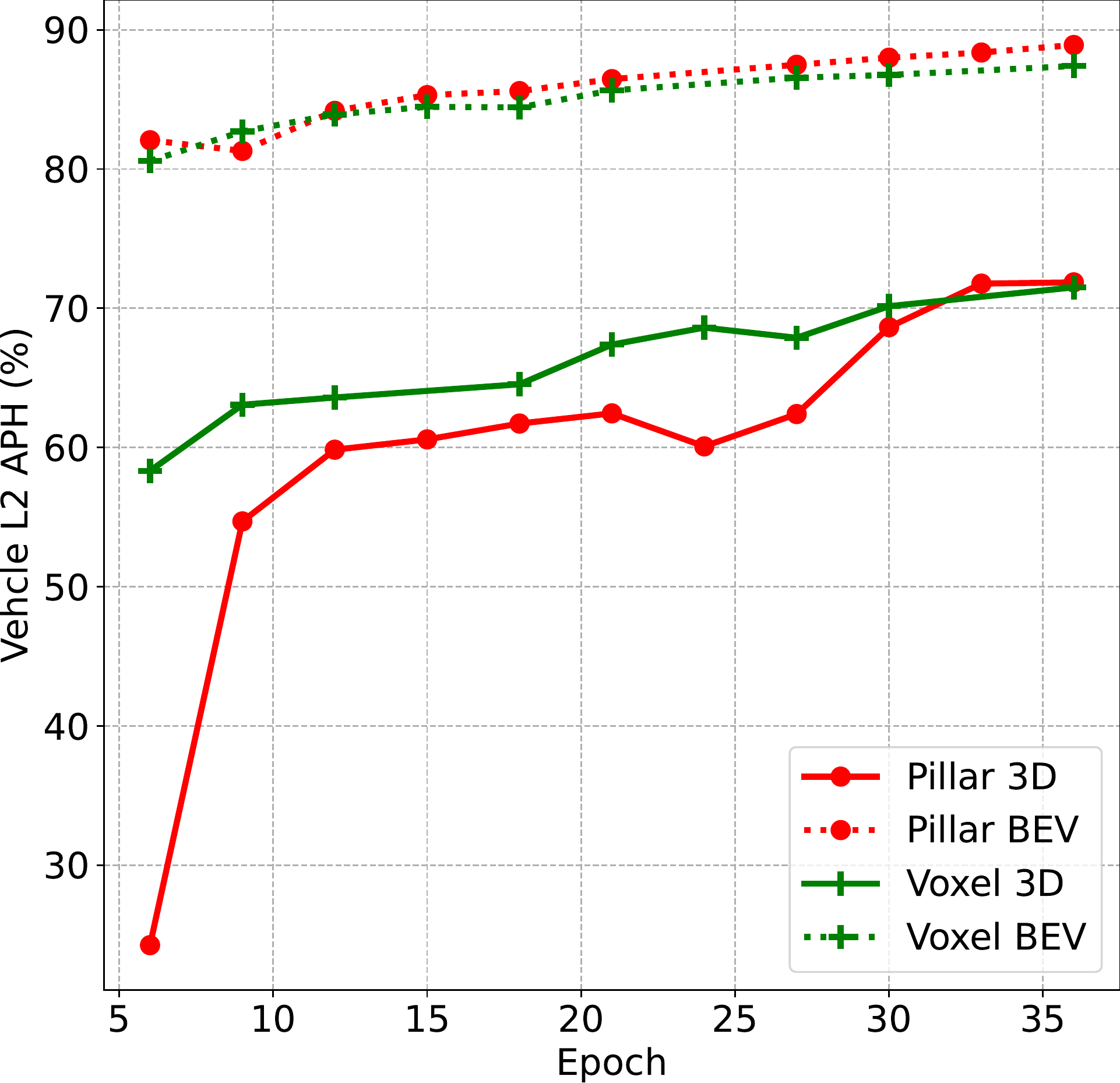}
   \vspace{-3mm}
   \caption{Comparison of the learning behaviors between the pillar and voxel based models. We report the L2 3D and BEV APH of vehicle on the validation set on WOD.}
   \label{fig:converage-vehicle}
   \vspace{-3mm}
\end{figure}

\subsection{Network Design Study}
\label{sec:study}

We perform extensive studies to analyze and understand the contribution of each individual network design. We first evaluate the impact of grid encoder, and demonstrate the importance of neck module, then investigate the effect of resolution, finally summarize the components one by one to show the improvement roadmap.   

\begin{table}[t]
\small
\tabcolsep=0.13cm
\centering
\begin{tabular}{lccc|c|c|c}
\hline
In Size & Backbone $\downarrow$ & Head $\uparrow$ & Out Size &  Veh & Ped & Latency \\ \hline
0.3 & 1 &  1 & 0.3 & 65.0 & 67.2 & 255 \\ 
0.075 & 8 & 1 & 0.6 &62.8 & 66.6 & 131 \\ 
0.075 & 8 & 2 & 0.3 & 64.8 & 69.0 & 173 \\ \hline
\end{tabular}
\caption{Comparison of different resolutions. We adopt the pillar size (m) to represent the resolutions of input grids and output features (consumed by head). We evaluate the overall downsampling rate in backbone and the upsampling rate in head. We report the L2 3D APH and latency (ms) on the validation set of WOD.} 
\label{tab:resolution}
\end{table}

\subsubsection{Study of Grid Encoders}
We begin with evaluating the three representative grid encoders as introduced in Section~\ref{grid-encoder}, i.e., pillar, voxel and MVF. Although the three encoders have been proposed for a long time, they have never been fairly compared under the same network architecture and grid resolution. Here, we experiment with a sparse ResNet-18~\cite{shi2022pillarnet} as the backbone for its effectiveness and efficiency. We scale the width of the backbone and obtain a series of networks ranging from tiny, small, base to large, namely Pillar/Voxel/MVF-T/S/B/L. Table~\ref{tab:capacity} lists the channel and parameter numbers, FLOPs, and latency of each model. Note there is no Voxel-T as FLOPs of voxel based models are much higher and the smallest one starts from a similar computational cost as Pillar-S.

For the first group in Table~\ref{tab:capacity}, we compare the three grid encoders with different model scales under the regular training schedule (i.e., 12 epochs), which is commonly adopted. As can be seen in the table, under BEV APH, the pillar encoder performs favorably on vehicle and is comparable on pedestrian, while with remarkably lower latency. This conclusion is further supported by the per-class comparison on nuScenes in Table~\ref{tab:nusc_class}. Note pedestrians usually take only a few pillars in the perspective of BEV, nevertheless, the pillar encoder is sufficient to achieve superior or on-par performance, including for the small objects.  

However, the pillar encoder still lacks behind under 3D APH. To further study the reasons for this gap, we enhance the models by extending the training schedule to 36 epochs with an extra IoU regression loss~\cite{shi2022pillarnet}, and incorporating an IoU score branch~\cite{hu2022afdetv2} in the multi-group detection head (one for vehicle and the other for pedestrian and cyclist, as illustrated in Figure~\ref{fig:sup-arch}).  
We call the models trained under this enhanced strategy as Pillar/Voxel/MVF+. As compared in the second group of Table~\ref{tab:capacity}, the pillar models achieve comparable or even better results than the voxel models on vehicle under 3D APH, while running considerably faster. We hypothesize that without explicit height modeling, the pillar based networks require refined designs such as longer training to be able to fully converge. This challenges the common belief that the pillar encoder loses height information, and suggests that the fine-grained local geometrical modeling may not be necessary.

\begin{figure}[t]
    \centering
    \includegraphics[width=\linewidth]{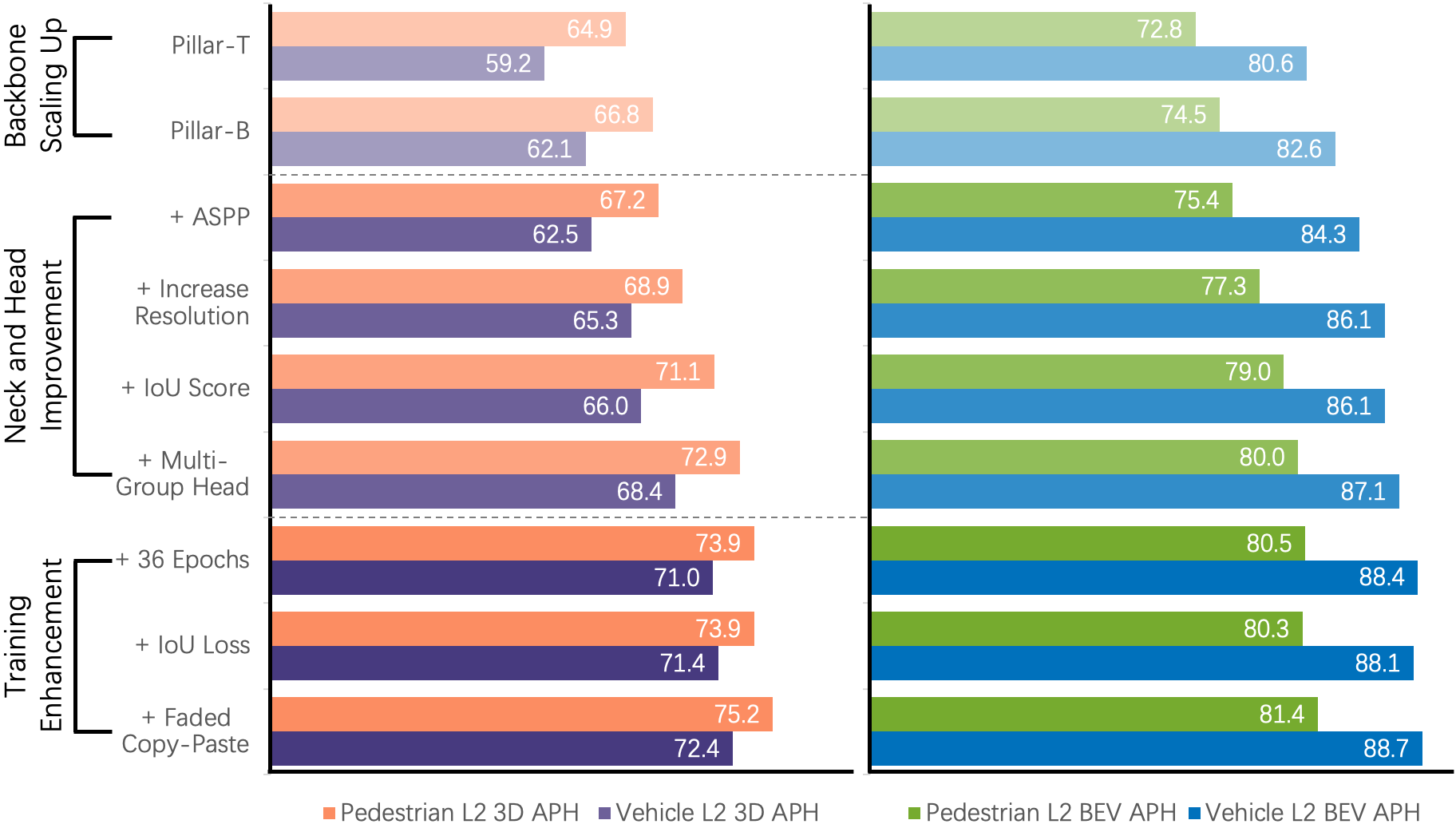}
    \caption{Improvement by each individual component to illustrate the performance boosting roadmap. We report the L2 3D and BEV APH of vehicle and pedestrian on the validation set of WOD.}
    \vspace{-2mm}
    \label{fig:roadmap}
\end{figure}

This counter-intuitive result motivates us to rethink how to efficiently and effectively allocate the computational resources for a 3D object detection network. The pillar based models allocate resources only in the BEV space, while the voxel and MVF based methods also spend computations along the height dimension. When comparing these methods, previous works fail to take into account the computational budget. In our experiments, we show that under similar FLOPs, allocating computations to the height dimension is not beneficial. We conclude that investing all FLOPs in the BEV space is not only more efficient but also more effective, as shown in Figure~\ref{fig:teaser} and Tables~\ref{tab:capacity} and~\ref{tab:nusc_class}.

It also reveals that training matters. Most previous works usually employ the regular or short training schedule for comparison, which could result in different conclusions. We demonstrate the learning behaviors of pillar and voxel based networks in Figure~\ref{fig:converage-vehicle}. Interestingly, the pillar model is found to converge comparably to the voxel model in BEV APH. While for 3D APH, the pillar model converges much slower than the voxel model. However, this gap diminishes when training continues for sufficient epochs, suggesting that the performance gap in 3D APH between pillar and voxel reported by the previous methods are partially caused by their different convergence rates, instead of the more fine-grained geometrical modeling in voxel.

\subsubsection{Study of Neck Modules}
In the previous study, we conclude that the different local point aggregators may not be essential to the final results. In the following, we show that a simple upgrade on the network architecture improves the performance greatly. In particular, we focus on the neck module design, which has not been well explored in 3D object detection.

Most current networks in the field rely on the multi-scale fusion as used in~\cite{zhou2018voxelnet, shi2022pillarnet}, which upsample feature maps from different stages to the same resolution and then concatenate them. How to design a neck module to conduct more effective feature aggregation has been extensively researched in 2D object detection, but most advanced techniques have not been adopted in 3D object detection. We first integrate with the two popular designs, i.e., FPN~\cite{lin2017feature} and BiFPN~\cite{tan2020efficientdet}. As shown in the first group of Table~\ref{tab:neck}, we observe up to 2.38\% improvement on vehicle over the neck developed in the most recent work PillarNet~\cite{shi2022pillarnet}.
 
One main goal of using multi-scale features is to deal with large variations of object scales. However, 3D objects in the BEV space do not suffer from such a problem. This motivates us to rethink whether the multi-scale representation is required for 3D object detection. We therefore investigate three single-scale neck modules. The baseline is a plain neck using a residual block without downsampling or upsampling, which gets inferior performance due to the limited receptive field. In YOLOF\cite{yolof}, it is argued that 2D object detector performs better when the receptive field matches with the object size. Inspired by this observation, we apply the dilated blocks as in YOLOF to enlarge the receptive field and yield better performance on vehicle. We also integrate with the ASPP block~\cite{Chen_2018_PAMI} and obtain up to 2.45\% improvement on vehicle compared with the neck used in PillarNet. 
All the above designs achieve comparable performance on pedestrian. These comparisons collectively imply that the multi-scale features may not be necessary, instead, enlarging receptive field plays the key role. 

This study demonstrates that simply adapting the neck modules from 2D object detection brings non-trivial improvements to 3D object detection, which is encouraging to explore more successful practices in the image domain to upgrade the network designs for point clouds.

\begin{table*}[]
\centering
\setlength{\tabcolsep}{4pt}
\begin{tabular}{l|l|cc|cc|cc|cc|cc|cc}
\hline
\multirow{2}{*}{Method} & \multirow{2}{*}{Frames} & \multicolumn{2}{c|}{Vehicle L1} & \multicolumn{2}{c|}{Vehicle L2} & \multicolumn{2}{c|}{Pedestrian L1} & \multicolumn{2}{c|}{Pedestrian L2} & \multicolumn{2}{c|}{Cyclist L1} & \multicolumn{2}{c}{Cyclist L2} \\  
                        &                         & AP            & APH           & AP            & APH           & AP           & APH         & AP           & APH         & AP           & APH         & AP           & APH         \\ \hline
SST-TS*~\cite{fan2022embracing} & 1 & 76.22&  75.79 &  68.04&  67.64&   81.39&  74.05 & 72.82&  65.93  &  - &  - & - & - \\
SWFormer~\cite{sun2022swformer} & 1 & 77.8 & 77.3  & 69.2 & 68.8  &  80.9 &  72.7  &  72.5 &  64.9  &  - &  - & - & - \\
PillarNet-18~\cite{shi2022pillarnet}   &  1    &    78.24 &  77.73  & 70.40 &  69.92  & 79.80  & 72.59  & 71.57  & 64.90  & 70.40  & 69.29 &  67.75  & 66.68       \\ 
AFDetV2~\cite{hu2022afdetv2} &  1   &  77.64  &   77.14  & 69.68  & 69.22 &80.19 & 74.62 & 72.16 & 66.95 & 73.72& 72.74 &71.06 & 70.12 \\ 
PV-RCNN++*~\cite{shi2021pv} & 1 & 79.25 & 78.78 & 70.61&  70.18 & 81.83 &  76.28 & 73.17 & 68.00 & 73.72 & 72.66 & 71.21 & 70.19 \\
PillarNeXt-B  &   1  &  78.40 & 77.90 & 70.27 & 69.81 & 82.53 &77.14 & 74.90 & 69.80 & 73.21 & 72.20 & 70.58 & 69.62 \\  
\hline
PillarNet-18~\cite{shi2022pillarnet}  &  2 &  79.59  &  79.06  & 71.56 &   71.08 &  82.11 &  78.82  & 74.49  & 71.35 & 70.41 &  69.57  &  68.27 & 67.46  \\
PillarNet-34~\cite{shi2022pillarnet}  & 2 & 79.98 & 79.47 & 72.00 & 71.53 & 82.52 & 79.33 & 75.00 & 71.95 & 70.51 & 69.69 & 68.38 & 67.58 \\ 
PV-RCNN++*~\cite{shi2021pv} & 2 & 80.17 & 79.70 & 72.14 & 71.70 & 83.48 & 80.42 & 75.54 & 72.61	& 74.63 & 73.75 & 72.35 & 71.50 \\ 
RSN*~\cite{sun2021rsn} & 3 & 78.4 & 78.1 &  69.5 & 69.1 & 79.4 & 76.2 & 69.9 & 67.0 & - & - & - & - \\ 
SST-TS*~\cite{fan2022embracing} & 3 & 78.66 & 78.21 & 69.98 & 69.57 & 83.81 & 80.14 & 75.94 & 72.37  & & & \\
SWFormer~\cite{sun2022swformer} & 3 & 79.4 & 78.9  & 71.1 & 70.6 &   82.9 & 79.0  & 74.8 &  71.1  & - & - & - & -  \\  
PillarNeXt-B &  3 & \textbf{80.58} & \textbf{80.08} & \textbf{72.89} & \textbf{72.42} & \textbf{85.04} & \textbf{82.11} & \textbf{78.04} & \textbf{75.19} &  \textbf{78.92} & \textbf{77.94} & \textbf{76.71} & \textbf{75.74} \\ \hline
CenterFormer~\cite{zhou2022centerformer} & 8 & 78.8  & 78.3   & 74.3  & 73.8  &  82.1  & 79.3   & 77.8  & 75.0   & 75.2  & 74.4   & 73.2  & 72.3 \\
MPPNet~\cite{chen2022mppnet} & 16 & 82.74 & 82.28 &  75.41 & 74.96  & 84.69 & 82.25  &  77.43 & 75.06  & 77.28 & 76.66 &  75.13 & 74.52  \\
3DAL$^\dagger$~\cite{qi2021offboard} & ALL  & 84.50 & - & - & - & 82.88 & - & - & -  & - & - & -  & -  \\ 
 \hline
\end{tabular}
\vspace{-2mm}
\caption{Comparison of PillarNeXt-B and the state-of-the-art methods under the 3D metrics on the validation set of WOD. * denotes the two-stage models and $^\dagger$ indicates the off-board method.}
\vspace{-1mm}
\label{tab:waymo-sota}
\end{table*}

\begin{table*}[t]
\centering
\setlength{\tabcolsep}{4pt}
\begin{tabular}{l|l|cc|cc|cc|cc|cc|cc}
\hline
\multirow{2}{*}{Method} & \multirow{2}{*}{Frames} & \multicolumn{2}{c|}{Vehicle L1} & \multicolumn{2}{c|}{Vehicle L2} & \multicolumn{2}{c|}{Pedestrian L1} & \multicolumn{2}{c|}{Pedestrian L2} & \multicolumn{2}{c|}{Cyclist L1} & \multicolumn{2}{c}{Cyclist L2} \\  
                        &                         & AP            & APH           & AP            & APH           & AP           & APH         & AP           & APH         & AP           & APH         & AP           & APH         \\ \hline
PV-RCNN++*~\cite{shi2021pv} & 1 & 91.57 & - & - & - & 85.43 & - & - & - & 75.94 & - & - & - \\ 
PillarNeXt-B  & 1 &93.30 & 92.60  & 87.26 & 86.53 & 88.19 & 82.13 & 81.77 & 75.82 & 75.67 & 74.61 & 72.97 & 71.95 \\  \hline
SWFormer~\cite{sun2022swformer} & 3 & 92.60 & - & - & -&  87.50 & - & - & - & - & - & - & - \\ 
PillarNeXt-B &  3 & \textbf{94.41}  & 93.73 & 89.36 & 88.66 & \textbf{90.20} & 86.94 & 84.66 & 81.36 & 81.35 & 80.32 & 79.23 & 78.22 \\ \hline
3DAL$^\dagger$~\cite{qi2021offboard} & ALL & 93.30 & - & - & -&  86.32 & - & - & - & - & - & - & - \\ \hline
\end{tabular}
\vspace{-2mm}
\caption{Comparison of PillarNeXt-B and the state-of-the-art methods under the BEV metrics on the validation set of WOD. * denotes the two-stage models and $^\dagger$ indicates the off-board method.}
\vspace{-2mm}
\label{exp:waymobev}
\end{table*}

\subsubsection{Study of Resolutions}\label{sec:resolution}
Intuitively, a smaller grid size retains more fine-grained information but requires a higher computational cost. Downsampling can effectively reduce the cost but degrade the performance. We experiment with different grid and feature resolutions by changing grid sizes and feature sampling rates in backbone and head. As shown in Table~\ref{tab:resolution}, if the output feature resolution is fixed (0.3), using a large grid size (0.075 to 0.3) does not affect the performance of large objects such as vehicles, but deteriorates the accuracy of small objects like pedestrians. Downsampling the output feature resolution (0.3 to 0.6) impairs the performance of both categories. However, if simply providing an upsampling layer in the detection head, we obtain significant improvement, especially for the small objects. This suggests that the fine-grained information may have already been encoded in the downsampled feature maps, and a simple upsampling layer in head can effectively recover the details.

\subsubsection{Summary}
We provide the improvement of each component one by one to elucidate the boosting roadmap in Figure~\ref{fig:roadmap}. As compared in this figure, one can see that the model scaling (e.g., tiny to base), the enhanced network neck and head (e.g., ASPP based neck and simple modifications in head), and the appropriate training (e.g., sufficient training epochs and data augmentation), produce tremendous improvements over the original baseline model. In the following experiments, we utilize Pillar-B with above improvements as the default setting for our proposed network PillarNeXt-B. It is extensively compared to the state-of-the-art methods that are specifically developed for point clouds. We illustrate the overall architecture of PillarNeXt in Figure~\ref{fig:sup-arch}.

\subsection{Comparison with State-of-the-Art on WOD}
We compare PillarNeXt-B with the published results on the validation set of WOD. As a common practice, we list the methods of using single and multiple frames separately. For completeness, we also compare to the methods with long-term temporal modeling. Our model is trained for 36 epochs with the faded copy-and-paste data augmentation.

As compared in Table~\ref{tab:waymo-sota}, our single-stage model outperforms many two-stage methods. It is also worth noticing that our pillar based approach without explicit temporal modeling~\cite{prernn} even achieves better results for small objects such as pedestrians than the methods with complex temporal modeling and fine-grained geometrical modeling. This clearly verifies the importance of network designs in terms of basic architecture and appropriate training.

\begin{table*}[t]
\centering
\setlength{\tabcolsep}{1.5pt}
\begin{tabular}{l|c|cc|cc|cc|cc|cc|cc|cc}
\hline
\multirow{2}{*}{Method} &  \multirow{2}{*}{Frames} &   \multicolumn{2}{c|}{All L2}  & \multicolumn{2}{c|}{Vehicle L1} & \multicolumn{2}{c|}{Vehicle L2} & \multicolumn{2}{c|}{Pedestrian L1} & \multicolumn{2}{c|}{Pedestrian L2} & \multicolumn{2}{c|}{Cyclist L1} & \multicolumn{2}{c}{Cyclist L2} \\  
                        &                  &  mAP & mAPH   & AP            & APH           & AP            & APH           & AP           & APH         & AP           & APH         & AP           & APH         & AP           & APH         \\ \hline
SWFormer~\cite{sun2022swformer}   & 3 & -  & - & 82.89   & 82.49 & 75.02& 74.65 & 82.13& 78.13 & 75.87& 72.07&  - & - & - & - \\ 
PillarNet-34$^\dagger$~\cite{shi2022pillarnet} &   3 &  73.98 & 72.48  &  83.23 &82.80 &76.09 &75.69 &82.38 &79.02 &76.66& 73.46 &71.44 &70.51 &69.20 &68.29 \\ 
CenterPoint++\cite{yin2021center} & 3  &  74.20  & 72.80 & 82.80 & 82.30 &  75.50 & 75.10  & 81.00 & 78.20 &  75.10 & 72.40 &  74.40 & 73.30  & 72.00 & 71.00 \\
AFDetV2~\cite{hu2022afdetv2}  & 2 &  74.60   & 73.12 & 81.65 & 81.22  & 74.30 & 73.89  & 81.26 & 78.05  & 75.47 & 72.41 &  76.41 & 75.37  & 74.05 & 73.04 \\ 
PV-RCNN++*~\cite{shi2021pv}   & 2  &  75.00  & 73.52  & 83.74 &83.32& 76.31 &75.92 &82.60& 79.38 &76.63 &73.55 &74.44 &73.43 &72.06 &71.09 \\ 
PillarNeXt-B  &  3  & \textbf{75.53}   & \textbf{74.10} & 83.28 & 82.83 & 76.18 & 75.76 & 84.40 & 81.44 &  78.84 & 75.98 & 73.77 & 72.73 & 71.56 & 70.55 \\ \hline
\end{tabular}
\vspace{-2mm}
\caption{Comparison of PillarNeXt-B and the state-of-the-art methods under the 3D metrics on the test set of WOD. * denotes the two-stage model and $^\dagger$ indicates using test-time augmentations.}
\vspace{-1mm}
\label{exp:waymotest}
\end{table*}

\begin{table*}[t]
\centering
\begin{tabular}{l|l|l|l|l|ccccc}
\hline
Method & Encoder & Grid Size & NDS & mAP & mATE$\downarrow$ & mASE$\downarrow$ & mAOE$\downarrow$ & mAVE$\downarrow$ & mAAE$\downarrow$   \\ \hline
 CenterPoint~\cite{yin2021center}  & V &  0.075  &  66.8    &   59.6  &  0.292 & 0.255 & 0.302 & 0.259 & 0.193    \\
 OHS~\cite{chen2020object} & V & 0.1 & 66.0 & 59.5  &  -  &  -  &  -  &  -  &  - \\
PillarNet-18~\cite{shi2022pillarnet} & P    &  0.075     &    67.4 &    59.9 &  -  &  -  &  -  &  -  &  -    \\ 
Transfusion-L~\cite{bai2022transfusion} & V  &  0.075 & 66.8 & 60.0  &  -  &  -  &  -  &  -  &  -    \\ 
UVTR-L~\cite{li2022unifying} & V  &  0.075  & 67.7  & 60.9  & 0.334  & 0.257   &0.300   &0.204  & 0.182 \\  VISTA~\cite{deng2022vista} & V+R & 0.1 & 68.1 & 60.8  &  -  &  -  &  -  &  -  &  -    \\ 
PillarNeXt-B  & P  &  0.075     &    \textbf{68.8} &   \textbf{62.5}  &  0.278 & 0.251 & 0.269 & 0.248 & 0.201\\ 
Our Voxel-B& V  & 0.075 & 68.2 & 62.4 & 0.278 & 0.250 & 0.308 & 0.263 & 0.198 \\ \hline
\end{tabular}
\vspace{-2mm}
\caption{Comparison of PillarNeXt-B and the state-of-the-art methods on the validation set of nuScenes. P/V/R denotes the pillar, voxel and range view based grid encoder, respectively. Most leading methods adopt the voxel based representations.} 
\vspace{-1mm}
\label{exp:nusc}
\end{table*}

\begin{table*}[h!]
\centering
\begin{tabular}{l|cccccccccc|c}
\hline
Method & Car & Truck & Bus & Trailer & CV & Ped & Motor & Bicycle & TC & Barrier & mAP\\ \hline
PillarNeXt-B & 84.8 & 58.6 & 66.5 & 35.3 & 21.4 & 87.2 & 68.0 & 56.4 & 77.0 & 69.8  & 62.5\\ 
Our Voxel-B & 84.3 & 58.3 & 69.3 & 37.1 & 21.4 & 87.4 & 67.6 & 54.7 & 75.0 & 69.2 & 62.4\\ \hline
\end{tabular}
\vspace{-2mm}
\caption{Comparison of our proposed pillar and voxel based models under per-class AP and mAP on the validation set of nuScenes. Abbreviations are construction vehicle (CV), pedestrian (Ped), motorcycle (Motor), and traffic cone (TC).}
\label{tab:nusc_class}
\end{table*}

In addition to 3D results, we also report BEV metrics in Table~\ref{exp:waymobev}. BEV representation is widely used in autonomous driving as the downstream tasks are naturally carried out in the space of BEV. Interestingly, our single-frame model already outperforms many multi-frame methods. As compared with the off-board method 3DAL~\cite{qi2021offboard}, which takes the whole sequence (around 200 frames) for refinement, our 3-frame model achieves better performance. This again validates the efficacy of our succinct single-stage network.

In Table~\ref{exp:waymotest}, we further demonstrate our results on the test set of WOD to evaluate the generalization of our approach. We do not use any test-time augmentation or model ensembling. PillarNeXt-B without bells and whistles is also found to outperform the state-of-the-art methods.

\subsection{Comparison with State-of-the-Art on nuScenes}
We in the end compare PillarNeXt-B with the state-of-the-art methods on nuScenes. Our model is trained for 20 epochs with the commonly used re-sampling CBGS~\cite{zhu2019class} and the faded copy-and-paste data augmentation. 
We report the results on the validation set in Table~\ref{exp:nusc}. Here we use a simpler model by removing the IoU score branch in the detection head. Our approach achieves the superior performance of 68.8\% NDS and 62.5\% mAP, which show the exceptional generalizability of the proposed model across different datasets. It is noteworthy that apart from PillarNet-18, all high-performing methods are voxel based. Our pillar based model outperforms the leading voxel and multi-view based methods by a large margin in mAP. For further analysis, we also compare with our voxel based model (Voxel-B) under exactly the same setting. PillarNeXt-B obtains higher NDS and comparable mAP. In particular, for the per-class performance in Table~\ref{tab:nusc_class}, PillarNeXt-B achieves on par or superior results compared to Voxel-B in pedestrians and traffic cones. This further verifies that the pillar based model can be highly effective in accurately detecting small objects.

\section{Conclusions}
In this paper, we challenge the common belief that a high-performing 3D object detection model requires the fine-grained local geometrical modeling. We systematically study three local point aggregators, and find that the simplest pillar encoder with enhanced strategy performs the best in both accuracy and latency. We demonstrate that enlarging receptive field and manipulating resolutions play the key roles. We hope our findings can serve as a solid baseline and encourage the research community to rethink what should be focused on for LiDAR based 3D object detection.

\newpage
{\small
\bibliographystyle{ieee_fullname}
\bibliography{egbib}
}

\newpage
\appendix
\section*{Appendix} 

\section{More Implementation Details}\label{sec:training_recipes}
\label{recipe}

We use 0.01 weight decay in the optimizer AdamW, and set the one-cycle scheduled learning rate to 0.001 when the batch size is 16. Our basic augmentations include random flipping, random rotation within the range of $[-\frac{\pi}{4}, \frac{\pi}{4}]$, random scaling between $[0.9, 1.1]$, and random translation with a noise factor of $0.5$, which are used for all experiments. Table~\ref{tab:aug} shows the detailed architecture and training settings used in various experiments. As can be seen in this table, for the experiments on WOD, ``Study'' denotes the regular setting in Section~\ref{sec:study}, and ``Study+'' indicates the enhanced version. Pillar/Voxel/MVF+ are trained with such enhanced strategy. We use the full settings with additional faded copy-and-paste augmentation when comparing with the single-frame or multi-frame based state-of-the-art methods.   

\begin{table*}[t]
\centering
\begin{tabular}{|l|cccc|c|}
\hline
\multirow{2}{*}{}                   & \multicolumn{4}{c|}{Waymo Open Dataset (WOD)}                                                                                                        & \multirow{2}{*}{nuScenes} \\ \cline{2-5}
                                    & \multicolumn{1}{l}{Study} & \multicolumn{1}{l}{Study+} & \multicolumn{1}{l}{Full (Single-Frame)} & Full (Multi-Frame) &                           \\ \hline    
Flip/Scaling/Rotation/Translation & $\checkmark$            & $\checkmark$      & $\checkmark$     &   $\checkmark$         &    $\checkmark$       \\ 
Random Drop Frames   & $\checkmark$    & $\checkmark$     &      &    $\checkmark$          &          \\ 
Faded Copy-and-Paste  &          &         & $\checkmark$        &   $\checkmark$       &     $\checkmark$          \\  \hdashline
IoU Regression Loss  &           & $\checkmark$            & $\checkmark$   & $\checkmark$     &    $\checkmark$   \\ 
IoU Score Branch     &           & $\checkmark$             & $\checkmark$      &     $\checkmark$     &                \\ 
Multi-Group Head & & $\checkmark$ & $\checkmark$ & $\checkmark$ &  $\checkmark$\\ \hdashline
Epochs   & 12        & 36         & 36               &   36       &     20       \\ \hline
\end{tabular}
\vspace{-2mm}
\caption{Details of the experimental settings for different experiments on the two benchmark datasets.}
\label{tab:aug}
\end{table*}

\end{document}